\documentclass[journal]{IEEEtran}

\ifCLASSINFOpdf
\usepackage[pdftex]{graphicx}
\graphicspath{{./}}

\DeclareGraphicsExtensions{.png,.pdf,.jpg,.jpeg}
\else
   \usepackage[dvips]{graphicx}
   \graphicspath{{Figures/}}
     \DeclareGraphicsExtensions{.png,.pdf,.jpg,.jpeg}
\fi

\usepackage{arydshln} 
\usepackage{fancyhdr}  

\usepackage{algorithmic}
\usepackage{algorithm}
\usepackage{array}
\usepackage[caption=false,font=normalsize,labelfont=sf,textfont=sf]{subfig}
\usepackage{textcomp}
\usepackage{stfloats}
\usepackage{url}
\usepackage{verbatim}
\usepackage{cite}
\usepackage{multirow}
\usepackage{pifont}
\usepackage{amsmath,amssymb,amsfonts,amsbsy}
\usepackage{bm} 
\usepackage{graphicx}
\usepackage{mathtools}
\usepackage{array,ragged2e}
\usepackage{booktabs}
\usepackage{makecell}

\usepackage{color}
\usepackage[table]{xcolor}
\definecolor{myPink}{rgb}{0.9294, 0.0078, 0.5490}
\definecolor{Gray}{gray}{0.92}
\definecolor{my_color}{HTML}{E8F3F1}
\definecolor{lightgray}{gray}{0.9}  
\usepackage[
  colorlinks=true,
  linkcolor=red,  
  citecolor=green,  
  urlcolor=myPink,   
  linkbordercolor={1 1 1} 
]{hyperref}
\usepackage{threeparttable}
\usepackage{diagbox}
\usepackage{orcidlink}

\makeatletter
\let\NAT@parse\undefined
\makeatother

\begin{document}

\title{MRC-DETR: An Adaptive Multi-Residual Coupled Transformer for Bare Board PCB Defect Detection} 


\author{
        Jiangzhong~Cao,
        Huanqi~Wu,
        Xu~Zhang, 
        Lianghong ~Tan,
        and Huan~Zhang


\thanks{This work was supported in part by the National Natural Science Foundation of China under Grant No. 62302105, and in part by the Guangdong Provincial Key Laboratory of Intellectual Property \& Big Data under Grant No. 2018B030322016. \emph{(Corresponding author: Huan Zhang.)}}%

\thanks{Jiangzhong Cao, Huanqi Wu, Lianghong Tan, and Huan Zhang are with the  School of Information Engineering, Guangdong University of Technology, Guangzhou 510006, China. (e-mail: cjz510@gdut.edu.cn; 2441025664@qq.com; 2112303085@mail2.gdut.edu.cn; huanzhang2021@gdut.edu.cn).}

\thanks{Xu Zhang is with the School of Computer Science, Wuhan University, Wuhan 430072, China (e-mail: zhangx0802@whu.edu.cn).}


}

\markboth{Journal of \LaTeX\ Class Files,~Vol.~14, No.~8, August~2021}%
{Shell \MakeLowercase{\textit{\textit{et al.}}}: A Sample Article Using IEEEtran.cls for IEEE Journals}

\maketitle

\begin{abstract} %
In modern electronic manufacturing, defect detection on Printed Circuit Boards (PCBs) plays a critical role in ensuring product yield and maintaining the reliability of downstream assembly processes. However, existing methods often suffer from limited feature representation, computational redundancy, and insufficient availability of high-quality training data—challenges that hinder their ability to meet industrial demands for both accuracy and efficiency. 
To address these limitations, we propose MRC-DETR, a novel and efficient detection framework tailored for bare PCB defect inspection, built upon the foundation of RT-DETR. 
Firstly, to enhance feature representation capability, we design a Multi-Residual Directional Coupled Block (MRDCB). This module improves channel-wise feature interaction through a multi-residual structure. Moreover, a cross-spatial learning strategy is integrated to capture fine-grained pixel-level relationships, further enriching the representational power of the extracted features.
Secondly, to reduce computational redundancy caused by inefficient cross-layer information fusion, we introduce an Adaptive Screening Pyramid Network (ASPN). This component dynamically filters and aggregates salient low-level features, selectively fusing them with high-level semantic features. By focusing on informative regions and suppressing redundant computations, ASPN significantly improves both efficiency and detection accuracy.
Finally, to tackle the issue of insufficient training data, particularly in the context of bare PCBs, we construct a new, high-quality dataset that fills a critical gap in current public resources. Our dataset not only supports the training and evaluation of our proposed framework but also serves as a valuable benchmark for future research in this domain.
Extensive experiments demonstrate that MRC-DETR achieving superior detection performance with reduced computational cost, thereby offering a more practical and reliable solution for real-world industrial applications. Our project page at https://github.com/utopiawsw/MRC-DETR.


\end{abstract}

\IEEEpeerreviewmaketitle

\section{Introduction}
\label{sec:intro} 
\IEEEPARstart{I}n modern electronic manufacturing, the quality of Printed Circuit Boards (PCBs) critically determines the performance and reliability of end products. As devices become increasingly compact and densely integrated, PCB defect detection \cite{PCBDetection_1,PCBDetection_2} faces growing technical challenges.
Although Automatic Optical Inspection (AOI) systems \cite{AOI_1,AOI_2} are widely used for large-scale screening, their limitations are becoming evident. First, existing AOI-based methods often fail to accurately detect micron-level defects \cite{Small_Object_Detection_1,Small_Object_Detection_2} or complex structures on high-density boards, resulting in frequent false alarms or missed detections. Second, AOI systems are highly sensitive to environmental conditions and process variations, leading to inconsistent detection performance across production lines. These issues not only degrade overall quality but also necessitate labor-intensive manual re-inspection, increasing costs and limiting efficiency.

Recent advances in deep learning offer promising solutions to these challenges. Unlike traditional image processing techniques that rely on manually designed features \cite{Traditional_image_processing}, deep learning-based object detection models \cite{SSD,Faster_R-CNN,CornerNet,YOLO,YOLOV2,YOLOV4} can automatically learn multi-level feature representations, demonstrating stronger robustness—especially in identifying subtle or irregular defects. This progress provides a viable path toward reducing both false positives and missed detections in AOI systems. However, for deep learning methods to be effectively deployed in industrial settings, several key challenges remain, including real-time inference, model stability, and adaptability to diverse production environments. Therefore, integrating the precision of deep learning with the practical requirements of PCB manufacturing is a critical direction for advancing intelligent quality control systems.

\begin{figure}[!tp]  
	\centerline{\includegraphics[page=1,trim = 0mm 0mm 0mm 0mm, clip, width=1\linewidth]{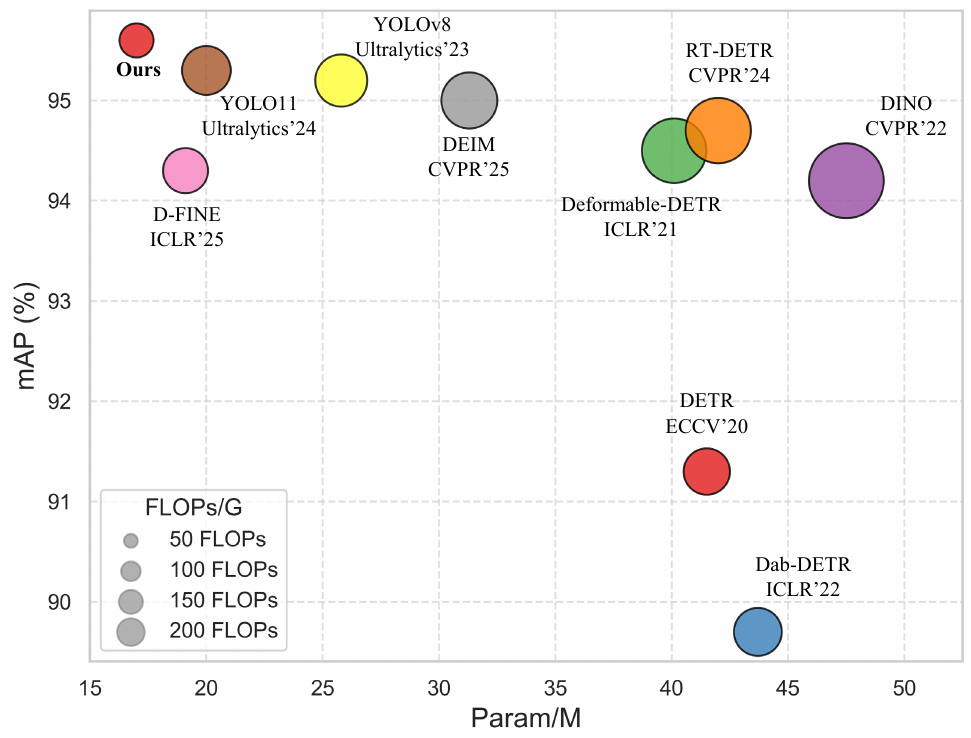}}
 \captionsetup{skip=0pt}
	\caption{Comparison of object detection models: mAP performance vs. parameter count (Param/M) and computational complexity (FLOPs/G).}
	\label{fig:motivation1}
	\vspace{-1em}
\end{figure}

The Detection Transformer (DETR) \cite{DETR} represents a paradigm shift in object detection by eliminating the need for anchor boxes and Non-Maximum Suppression (NMS) \cite{NMS} post-processing, which are commonly used in traditional methods such as YOLO . Instead, DETR formulates object detection as a set prediction problem \cite{end-to-end}, enabling truly end-to-end learning. Thanks to its global attention mechanism, DETR achieves strong detection accuracy and excels at modeling long-range dependencies, making it particularly suitable for detecting defects in complex scenes. However, DETR suffers from high computational complexity and slow inference speed, which limit its applicability in real-time industrial inspection tasks such as PCB defect detection \cite{Defect_Detection_survey}. 

To address these limitations, RT-DETR \cite{RT-DETR} was proposed as an improved variant that optimizes both the self-attention mechanism and detection head design, achieving significantly faster inference while maintaining competitive accuracy. As a result, RT-DETR strikes a favorable balance between detection performance and speed, making it a promising solution for real-time industrial applications.
Despite its success in various vision tasks, applying RT-DETR to PCB defect detection still faces problems such as insufficient feature extraction and a surge in computational complexity caused by multi-scale feature fusion \cite{FPN, Perceive-IR, PCB_defect1, Unveiling_the_underwater_world, PCB_defect2}. These issues motivate the need for further improvements in both efficiency and feature representation tailored to industrial PCB inspection scenarios.

To this end, we propose an improved RT-DETR-based method to enhance the performance of PCB defect detection. To demonstrate the efficiency and effectiveness of our proposed method, we compare it with several representative object detection models in terms of mean Average Precision (mAP), parameter count, and computational complexity. Fig. \ref{fig:motivation1} illustrates this comparison, showing that our method achieves competitive accuracy with significantly lower computational cost, making it well-suited for deployment in resource-constrained industrial settings. First, to mitigate the issues of insufficient feature representation and computational redundancy \cite{EMA,Faster_block}, we design the Multi-Residuals Directional Coupled Block (MRDCB). This module enhances feature expressiveness through parallel sub-networks and cross-space learning, while leveraging a multi-residual structure to reduce computational overhead without sacrificing representational capacity.
Second, to prevent the surge in computational complexity caused by cross-layer information interaction \cite{HS_FPN,ELA}, we introduce the Adaptive Screening Pyramid Network (ASPN). ASPN incorporates a lightweight spatial screening mechanism that enables efficient multi-scale feature fusion while maintaining low computational cost.
Finally, considering the lack of publicly available datasets specifically targeting bare PCB boards, we construct a new, high-quality dataset tailored for this stage of inspection. This dataset fills a critical gap in existing resources and provides valuable support for training and evaluating deep learning-based defect detection methods.

In summary, our main contributions are as follows:  

\noindent \ding{113}~We propose a Multi-Residuals Directional Coupled Block that enhances channel-wise feature interaction through a grouped multi-residual structure, along with a cross-space learning strategy to capture pixel-level relationships. This design improves detection accuracy while maintaining low computational cost, effectively addressing issues of limited feature representation and redundancy in PCB defect detection.

\noindent \ding{113}~To reduce the computational burden caused by cross-layer feature fusion, we introduce an Adaptive Screening Pyramid Network with a lightweight spatial screening module. It efficiently identifies and fuses salient low-level features with high-level semantics, enhancing multi-scale perception without increasing complexity.

\noindent \ding{113}~Recognizing the lack of public datasets for bare PCB defect detection, we construct a new, industry-relevant dataset that fills this gap and provides valuable data support for training and evaluating deep learning-based methods in real-world industrial settings.

\section{Related Work}
\subsection{Adaptive Feature Learning for PCB Defect Detection} 


In PCB defect detection tasks, due to the diversity of manufacturing processes, board complexity, and defect morphology, the model must have the ability to dynamically adapt to input features in order to effectively extract discriminative semantic representations. Traditional manual feature methods are difficult to adapt to complex backgrounds and differences in different types of defects. The rise of deep learning has driven the shift from static features to data-driven, self-adjusting feature modeling mechanisms, but existing methods still face challenges in multi-scale response, regional attention, and resource adaptation, especially on edge devices required for industrial deployment.


Recent studies have increasingly focused on enhancing the adaptability of feature learning to address the diverse and complex characteristics of PCB defects. Tang et al. \cite{Feature_extraction_1} introduced an autoencoder-based framework with low-rank constraints and Laplacian operators, allowing the model to adaptively suppress redundant information and highlight discriminative features. Yang et al. \cite{Feature_extraction_2} proposed a Multi-Receptive Field Block capable of adaptively capturing scale-variant patterns through a parameter-efficient design. 
Building upon these efforts, Yu et al.  \cite{Feature_extraction_4} designed a Double-Gated Convolution Extraction Block that dynamically adjusts feature selection for efficient learning, even on edge devices with constrained resources. 

The common point of these studies is the use of adaptive receptive fields, dynamic feature fusion, and attention mechanisms to enable the model to autonomously identify information-rich areas based on input features and enhance the ability to respond to key patterns. To ensure practical deployment, lightweight techniques such as model pruning  \cite{Model_pruning}, knowledge distillation  \cite{Knowledge_Distillers}, and edge computing \cite{Edge_computing} are increasingly being integrated into the adaptive feature learning framework. These methods not only improve the robustness and generalization of detection, but also lay the foundation for the practical deployment of adaptive feature learning in real-time industrial detection.

\subsection{Feature Interaction and Fusion for PCB Defect Detection} 


In PCB defect detection, the defect morphology has cross-scale and multi-structure characteristics, and relying on a single-level feature is often not enough to achieve accurate identification. Therefore, how to effectively integrate information of different scales and semantic levels and construct a robust feature representation has become a key research direction in recent years.


Researchers have proposed a series of efficient feature fusion strategies to meet this challenge. For example, Wang et al. \cite{Feature_fusion_1} proposed a pyramid convolution architecture based on 3D feature alignment, which enhances multi-scale representation capabilities while reducing redundant calculations; the Image Pyramid Guided Network (IPG-Net) proposed by Liu et al. \cite{Feature_fusion_2} introduces an auxiliary pyramid path to enrich spatial details, and improves the effect of small defect detection through feature alignment mechanism; in addition, another idea is to build a cross-layer interaction mechanism, such as the bidirectional feature interaction module proposed by Wang et al. \cite{Feature_fusion_3}, so that high-level semantics and low-level details can be mutually enhanced, thereby improving the detection performance under cross-resolution. In order to further combine local details with global context, Zheng et al. \cite{Feature_fusion_4} designed the Spatial and Channel-Enhanced Self-Attention (SC-SA) module, which combines local convolution with global self-attention mechanism to achieve joint enhancement in space and channels. These structures not only optimize the feature fusion path but also enhance the effective flow of information between different scales.
Recent research has shifted from conventional structure stacking strategies toward more efficient information flow-guided paradigms. Techniques such as the variable information bottleneck \cite{Variational_Information_Bottleneck,UniUIR,PCB_defect3, ESWA,PCB_defect4,Feature_extraction_3} have been introduced to actively suppress redundant information while preserving discriminative features during feature fusion. These adaptive strategies improved the robustness of the model in challenging scenarios, including complex backgrounds, low-contrast defects, and large variations in the size of the defect. 


\subsection{PCB Dataset}

In PCB defect detection research, the quality and diversity of datasets play a critical role in both model training and performance evaluation. An ideal dataset should encompass a wide range of defect types, various PCB layouts, and diverse environmental conditions such as lighting variations and background complexity.
Currently, publicly available PCB defect datasets are limited in scale and scope. The most widely used dataset was released by Peking University, containing 1,386 images with annotations for six types of defects, supporting tasks such as detection, classification, and registration \cite{Beida_Dataset}. However, this dataset focuses on finished PCB boards, with relatively simple backgrounds and limited scene variability. Another notable dataset is DeepPCB proposed by Tang et al. \cite{DeepPCB}, which includes 1,500 image pairs (template and test images) covering six defect categories. The dataset primarily consists of binary images with synthetically generated defects, which may not fully reflect real-world scenarios.

Although existing datasets have made contributions in this field, they mainly focus on assembled PCBs, lack coverage of bare board defects, have limited PCB design diversity, and have a single environment setting. This limits the generalization and robustness of detection models in actual industrial applications. In order to make up for the limitations of existing datasets in terms of defect types, board structures, and scene diversity, we constructed a set of PCB defect detection datasets for the bare board stage. This dataset is collected from actual AOI production lines and covers typical defect types such as open circuits, short circuits, and hole position deviations. All images are manually cropped and finely labeled to ensure the accuracy and consistency of the annotations. Compared with existing synthetic datasets, this dataset is closer to actual application scenarios, can provide high-quality support for detection models with stronger robustness and generalization capabilities, and help promote the actual deployment of PCB defect detection.

\begin{figure*}[!htp]  
	\centerline{\includegraphics[page=1,trim = 0mm 0mm 0mm 0mm, clip, width=0.98\linewidth]{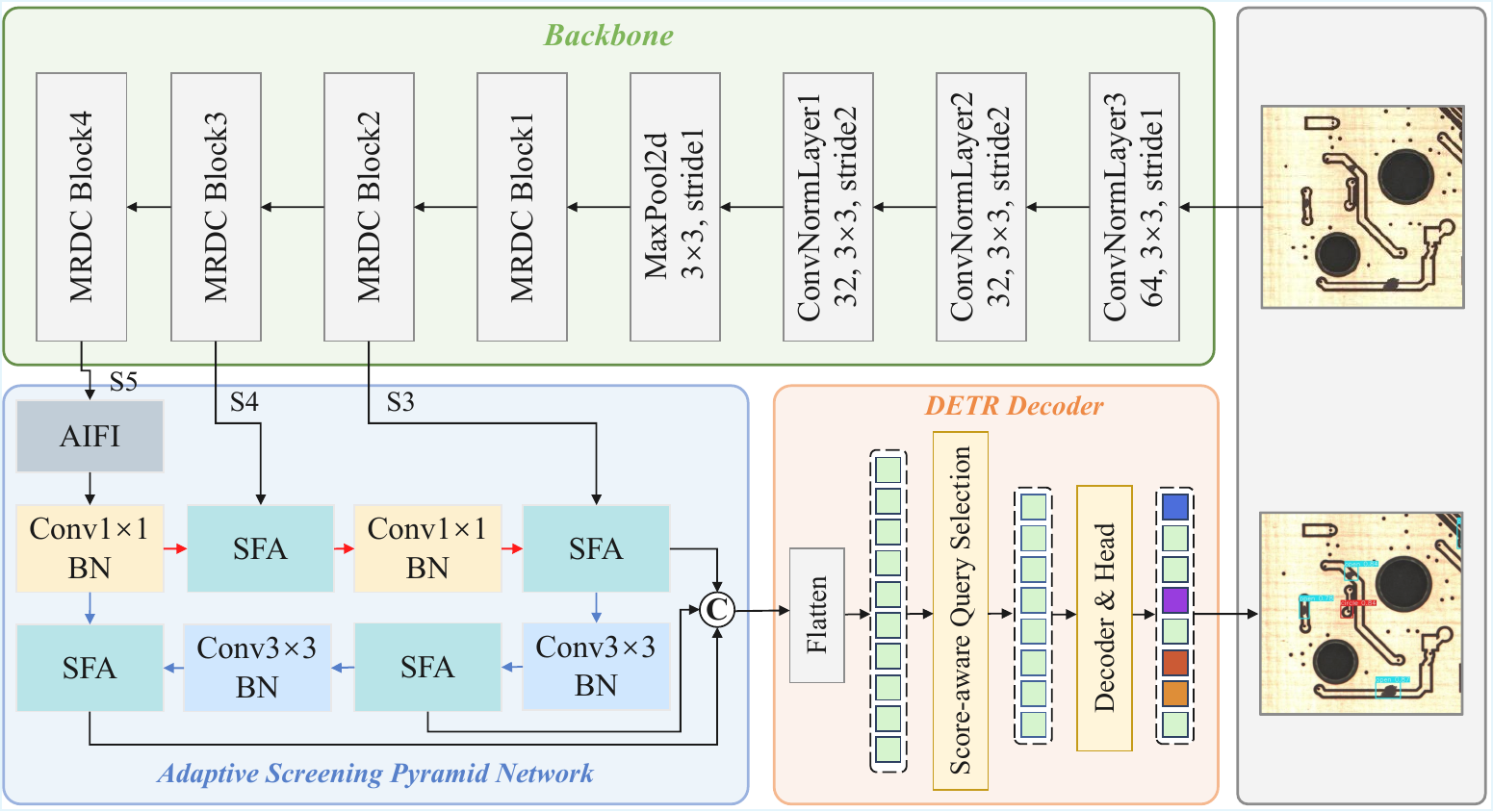}}
  \captionsetup{skip=0pt}
	\caption{Overview of the proposed MRC-DETR. The Multi-Residuals Directional Coupled (MRDC) Block outputs multiscale features S3, S4, S5. The Adaptive Screening Pyramid Network converts the multiscale features into a
sequence of image features using a Selective Feature Aggregation (SFA) and Attention-based Intra-scale Feature Interaction (AIFI). AIFI is inherited from RT-DETR, and it enhances intra-scale feature representation by capturing long-range dependencies within each scale. Score-aware query selection is used to select a fixed number of image features as initial object queries for the decoder. Finally, the decoder with an auxiliary prediction header performs iterative optimization of the object query to generate boxes and confidence scores.}  
	\label{fig:MRC-DETR}
	\vspace{-1.5em}
\end{figure*}

\section{Method}
\label{Method}

\subsection{Overview Architecture } 
We propose an improved Transformer object detection framework, MRC-DETR.  As illustrated in Fig. \ref{fig:MRC-DETR}, for PCB bare board defect detection tasks. This method is based on RT-DETR with ResNet-50 as the backbone network, and optimizes the original model in a targeted manner, considering that PCB defects are usually small and sparse. The original model has two major problems: \textbf{1)} The backbone network has large computational redundancy and lacks sufficient adaptive feature learning capabilities; \textbf{2)} The fusion of low-level and high-level semantic features is insufficient, which limits the model's ability to express tiny defects.

To address these issues, we introduce two key components. First, the Multi-Residuals Directional Coupled Block (MRDCB) is designed to replace the original heavy backbone structure, significantly reducing computational cost while enhancing the model's capacity for fine-grained feature extraction and adaptive channel interaction. Second, we design the Adaptive Screening Pyramid Network (ASPN) to improve cross-scale information fusion. ASPN leverages high-level semantic features as spatial priors to adaptively filter low-level features, followed by a pixel-wise addition to fuse the refined low-level and high-level features. This selective fusion strategy enhances the network's ability to localize and represent subtle defects, ultimately improving overall detection performance on bare board PCB images.

\subsection{Multi-Residuals Directional Coupled Block } 
PCB defects usually only occupy a small area in the image and are easily submerged by background features in deep neural networks, resulting in the loss of key information. In addition, the high computational complexity of deep networks also limits their deployment and application in actual industrial scenarios.

\begin{figure}[!tp]  
	\centerline{\includegraphics[page=1,trim = 0mm 0mm 0mm 0mm, clip, width=1\linewidth]{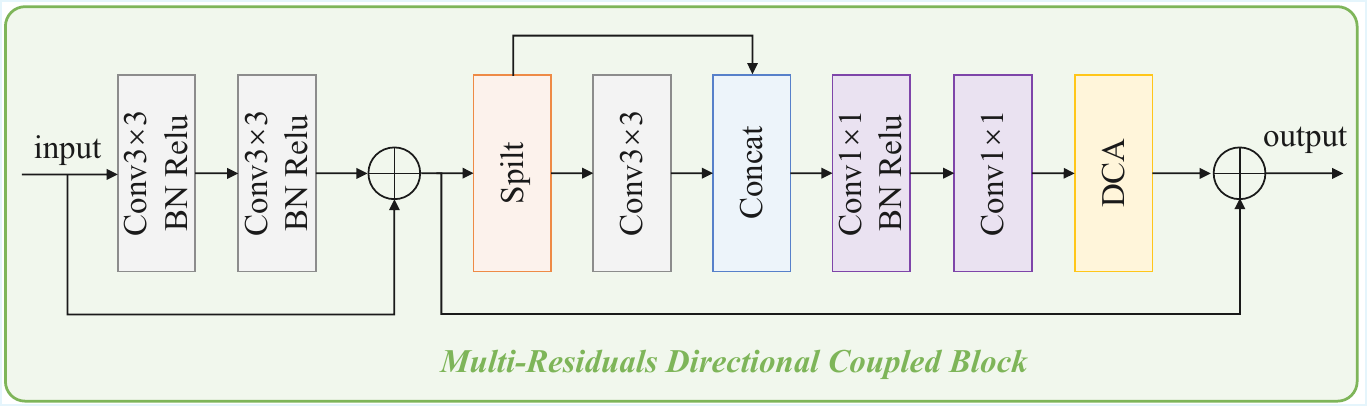}}
  \captionsetup{skip=0pt}
	\caption{Overview of the Multi-Residuals Directional Coupled Block (MRDCB), comprising a Multi-scale  Residual Unit and Directional Coupled Attention for efficient and enhanced feature representation.}  
	\label{fig:MRDCB}
	\vspace{-1.5em}
\end{figure}

To address these problems, we proposed the MRDCB. As illustrated in Fig.  \ref{fig:MRDCB},  which aims to enhance the model's adaptive feature learning capabilities while reducing computational overhead and improving the accuracy and efficiency of defect detection. The MRDCB consists of two key parts: Multi Scale Residual Unit and Directional Coupled Attention. The former retains and refines fine-grained features through a multi-level information transfer mechanism, while the latter constructs spatially aware and semantically related attention distributions through grouped channel interactions, thereby achieving lightweight and effective feature enhancement.

\subsubsection{Multi Scale Residual Unit} 
In deep neural networks, as the number of layers increases, PCB defect features are easily overwhelmed by background information, leading to a decline in detection performance. To address this issue, we propose a multi-scale residual connection mechanism that integrates feature splitting, channel interaction, and attention enhancement, ensuring that critical feature information is preserved and effectively propagated through the network. The first-scale residual connection directly links the input and output of the initial feature extraction module, retaining original feature information at an early stage. Building upon this, the second-level residual connection introduces channel interaction and an attention mechanism, which strengthen key feature representation across different abstraction levels while maintaining computational efficiency. Specifically, the architecture begins with two standard 3$\times$3 convolutional blocks for initial feature extraction, followed by the implementation of the first-level residual connection: 
\begin{equation}
\boldsymbol{\mathrm{R}}_\mathrm{1}=\boldsymbol{\mathrm{X}}+C B R_{2}\left(C B R_{1}(\boldsymbol{\mathrm{X}}
)\right),
\end{equation}
where $\boldsymbol{\mathrm{X}}
$ represents the input features, $\boldsymbol{\mathrm{R}}_\mathrm{1}$ represents the output of the first layer residual connection, and ${CBR}$ represents the convolution-normalization-activation function structure.


This direct residual connection can effectively alleviate the problem of gradient vanishing and enhance the network’s utilization of low-level features to a certain extent, enabling more efficient feature propagation through the deep architecture. Following this, we perform feature splitting on $\boldsymbol{\mathrm{R}}_\mathrm{1}$, dividing it into two parts: the first part, $\boldsymbol{\mathrm{S}}_\mathrm{1}$, is directly fed into an additional convolutional layer to extract new high-level features. The second part, $\boldsymbol{\mathrm{S}}_\mathrm{2}$, is concatenated with the output of the aforementioned convolutional layer, allowing for effective fusion of features across different levels. The resulting concatenated feature map can be expressed as follows:
\begin{equation}
\boldsymbol{\mathrm{F}}_\mathrm{split}=\mathrm{Concat}(Conv_{3\times3}(\boldsymbol{\mathrm{S}}_\mathrm{1}),\boldsymbol{\mathrm{S}}_\mathrm{2}).
\end{equation}

In order to further strengthen the key features, we use two 1$\times$1 convolutions for channel interaction based on the first layer of residual connections:\begin{equation}
\boldsymbol{\mathrm{F}}_\mathrm{interact}=Conv_{1\times1}(Conv_{1\times1}(\boldsymbol{\mathrm{F}}_\mathrm{split})),
\end{equation}
where the first 1$\times$1 convolution is used to expand the number of channels, increase the network width, and enhance the feature expression ability. The second 1$\times$1 convolution is used to adjust the channel dimension and enhance the information interaction between different channels.

Finally, we propose a Directional Coupled Attention (DCA) module, which computes channel-wise attention to refine key features. The refined output is then fused with the first-layer residual result  $\boldsymbol{\mathrm{R}}_\mathrm{1}$ via an additional residual connection. \begin{equation}
\boldsymbol{\mathrm{R}}_\mathrm{1}=\boldsymbol{\mathrm{R}}_\mathrm{1}+\mathrm{DCA}(\boldsymbol{\mathrm{F}}_{\mathrm{interact}}).
\end{equation}

Through the above operations, the network can retain key defect features at different depths, while reducing the amount of calculation and improving detection accuracy. Compared with ordinary single residual connections, the advantage of multi-scale  residual connections is that they enhance feature extraction capabilities while reducing the amount of calculation. Through a multi-layer residual structure, it can effectively prevent PCB defect features from disappearing during the network deepening process and maintain multi-scale information. The first layer of residuals fuses features at different levels to enhance the expression of local details. The second layer of residuals combines 1$\times$1 convolution and attention mechanisms to improve information interaction between channels, allowing the network to pay more attention to key defect areas. Overall, this design not only improves detection accuracy, but also reduces computational overhead, improving the efficiency and generalization ability of the model.

\subsubsection{Directional Coupled Attention} 
\begin{figure}[!tp]  
	\centerline{\includegraphics[page=1,trim = 0mm 0mm 0mm 0mm, clip, width=1\linewidth]{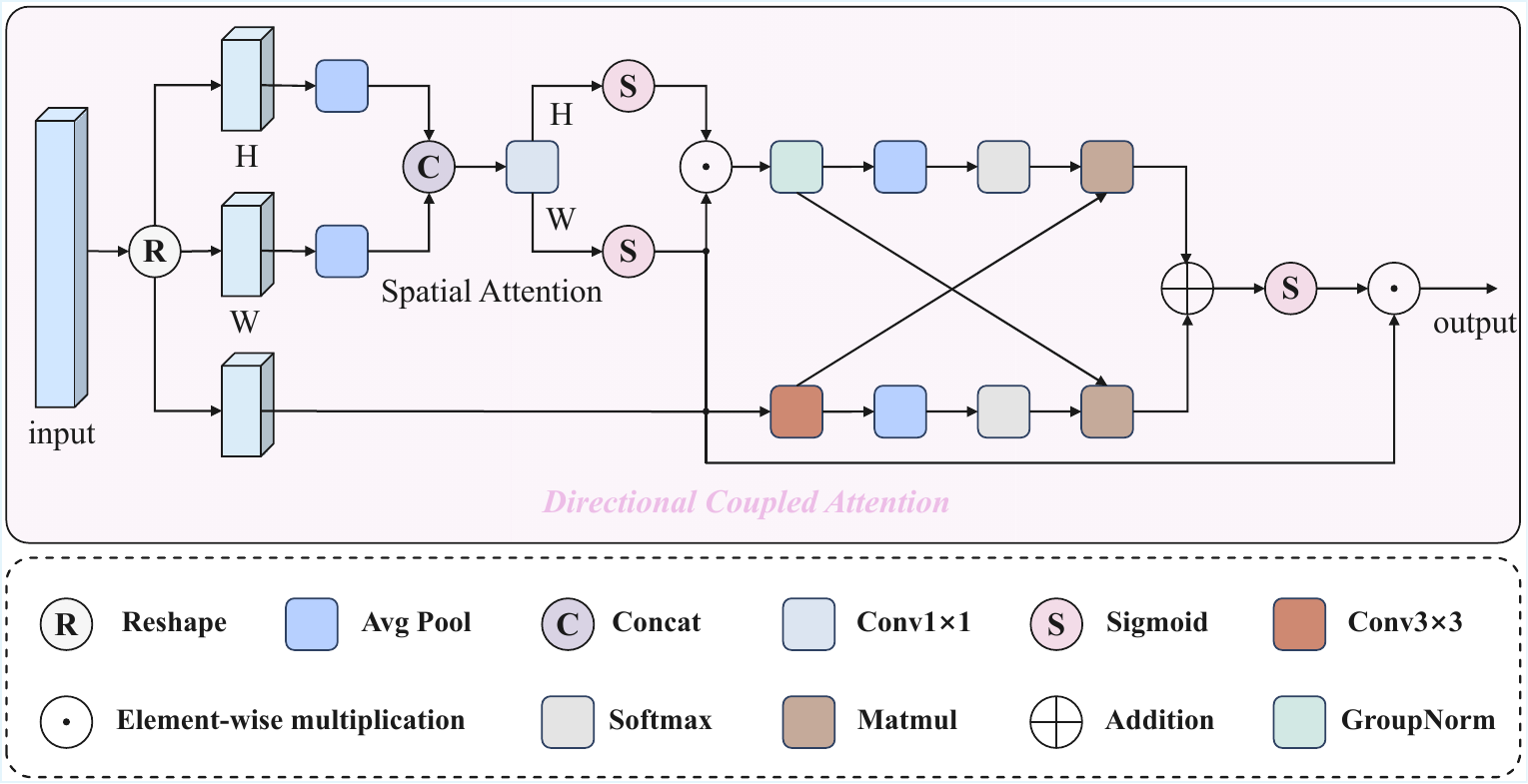}}
  \captionsetup{skip=0pt}
	\caption{Structure of the Direction-aware Coupled Attention (DCA). The DCA module is designed to enhance directional sensitivity and local detail retention in PCB defect detection.}  
	\label{fig:DCA}
	\vspace{-1.5em}
\end{figure}

As illustrated in Fig. \ref{fig:DCA}, DCA mainly consists of four steps: feature grouping, direction-aware feature extraction, parallel feature enhancement, attention calculation and feature weighting. Each part works together to improve the feature expression ability. The input feature map $\boldsymbol{\mathrm{X}}
\in\mathbb{R}^{C\times H\times W}$ is re-divided into multiple subgroups, and the number of channels in each subgroup is reduced, thereby reducing the computational overhead:
\begin{equation}
\boldsymbol{\mathrm{X}}_\mathrm{grouped}
=\mathrm{Reshape}(\boldsymbol{\mathrm{X}}
, C/G, H, W).
\end{equation}

In the PCB defect detection task, many defects are distributed along a specific direction. For example, short circuit defects may appear as long and thin strips, while hole deviation defects may be concentrated in certain local areas. Therefore, it is difficult to effectively model these directional feature distributions by relying solely on traditional convolution operations. In order to enhance the model's perception of directional information, DCA uses horizontal pooling and vertical pooling to enhance the direction perception of input features. The pooled features are mapped by 1$\times$1 convolution to further integrate features in different directions:
\begin{equation}
\boldsymbol{\mathrm{X}}_{\mathrm{W}}
=\mathrm{AvgPool}_{w}(\boldsymbol{\mathrm{X}}_\mathrm{grouped}
),
\end{equation}
\begin{equation}
 \boldsymbol{\mathrm{X}}_{\mathrm{H}}
=\mathrm{AvgPool}_{h}(\boldsymbol{\mathrm{X}}_\mathrm{grouped}
),
\end{equation}
\begin{equation}
\boldsymbol{\mathrm{X}}_{\mathrm{HW}}
=Conv_{1\times1}(\mathrm{Concat}(\boldsymbol{\mathrm{X}}_{\mathrm{H}}
,\boldsymbol{\mathrm{X}}_{\mathrm{W}}
)),
\end{equation}
where $\mathrm{AvgPool}_{w}$ and $\mathrm{AvgPool}_{h}$ denote average pooling operations applied along the height and width dimensions, respectively, aiming to capture global contextual information in each spatial direction.  The pooled features are concatenated and fused via a $1\times1$ convolution to produce the attention   features $\boldsymbol{\mathrm{X}}_{\mathrm{HW}}$. Then,  $\boldsymbol{\mathrm{X}}_{\mathrm{HW}}$ is split back into $\boldsymbol{\mathrm{X}}_{\mathrm{H}}^{\prime}$ and  $\boldsymbol{\mathrm{X}}_{\mathrm{W}}^{\prime}
$  and normalized using Sigmoid to ensure that the attention weights remain in a reasonable range, and then the original input is enhanced using direction-aware features:
\begin{equation}
\boldsymbol{\mathrm{X}}_\mathrm{1}
=\boldsymbol{\mathrm{X}}_\mathrm{grouped}
\odot\sigma(\boldsymbol{\mathrm{X}}_{\mathrm{H}}^{\prime})\odot\sigma(\boldsymbol{\mathrm{X}}_{\mathrm{W}}^{\prime})^T.
\end{equation}

Although direction-aware pooling can effectively extract global structural information, local detail information of defects is also crucial, especially the edge contours of small defects may be weakened in the pooling process. To compensate for this shortcoming, we introduce a parallel convolution feature enhancement channel to maintain local spatial information through 3$\times$3 convolution operations:
\begin{equation}
\boldsymbol{\mathrm{X}}_\mathrm{2}
={Conv}_{3\times3}(\boldsymbol{\mathrm{X}}_\mathrm{grouped}
).
\end{equation}

In order to enhance the information interaction between channels, we introduce a Softmax mechanism based on global pooling to calculate the attention weights so as to adaptively adjust the importance of feature channels. Global pooling is performed on the two feature channels respectively, and the calculated weight matrices $\boldsymbol{\mathrm{W}}_\mathrm{1}$ and $\boldsymbol{\mathrm{W}}_\mathrm{2}$ are applied to the features of the two branches, and then feature fusion is performed to obtain $\boldsymbol{\mathrm{X}}_\mathrm{1,2}$, and finally the original channel input is restored to $\boldsymbol{\mathrm{X}}_{{output}}\in\mathbb{R}^{C\times H\times W}$:
\begin{equation}
\boldsymbol{\mathrm{W}}_\mathrm{1}=\mathrm{Softmax}(\mathrm{AvgPool}(\mathrm{GroupNorm}(\boldsymbol{\mathrm{X}}_\mathrm{1}
))),
\end{equation}
\begin{equation}
\boldsymbol{\mathrm{W}}_\mathrm{2}=\mathrm{Softmax}(\mathrm{AvgPool}(\mathrm{GroupNorm}(\boldsymbol{\mathrm{X}}_\mathrm{2}
))),
\end{equation}
\begin{equation}
\boldsymbol{\mathrm{X}}_\mathrm{1,2}
=\boldsymbol{\mathrm{X}}_\mathrm{grouped}
\odot\sigma((\boldsymbol{\mathrm{W}}_\mathrm{1}\otimes\boldsymbol{\mathrm{X}}_\mathrm{2}
)+(\boldsymbol{\mathrm{W}}_\mathrm{2}\otimes\boldsymbol{\mathrm{X}}_\mathrm{1}
)).
\end{equation}

Through the above operations, the model's perception of PCB defect areas is improved while reducing computational redundancy. First, the input features are divided into multiple channel groups to reduce computational complexity. Then, global pooling is performed in the horizontal and vertical directions to extract spatial structure information, and directional features are fused through 1$\times$1 convolution. Next, the residual enhancement module and 3$\times$3 convolution are used to further enhance the local feature expression, and adaptive attention weights are constructed to enhance the feature response of the defective area. Finally, the input features are adjusted based on channel-level and spatial-level weights to improve the accuracy and robustness of defect detection.

\subsection{Adaptive Screening Pyramid Network} 
\begin{figure}[!tp]  
	\centerline{\includegraphics[page=1,trim = 0mm 0mm 0mm 0mm, clip, width=1\linewidth]{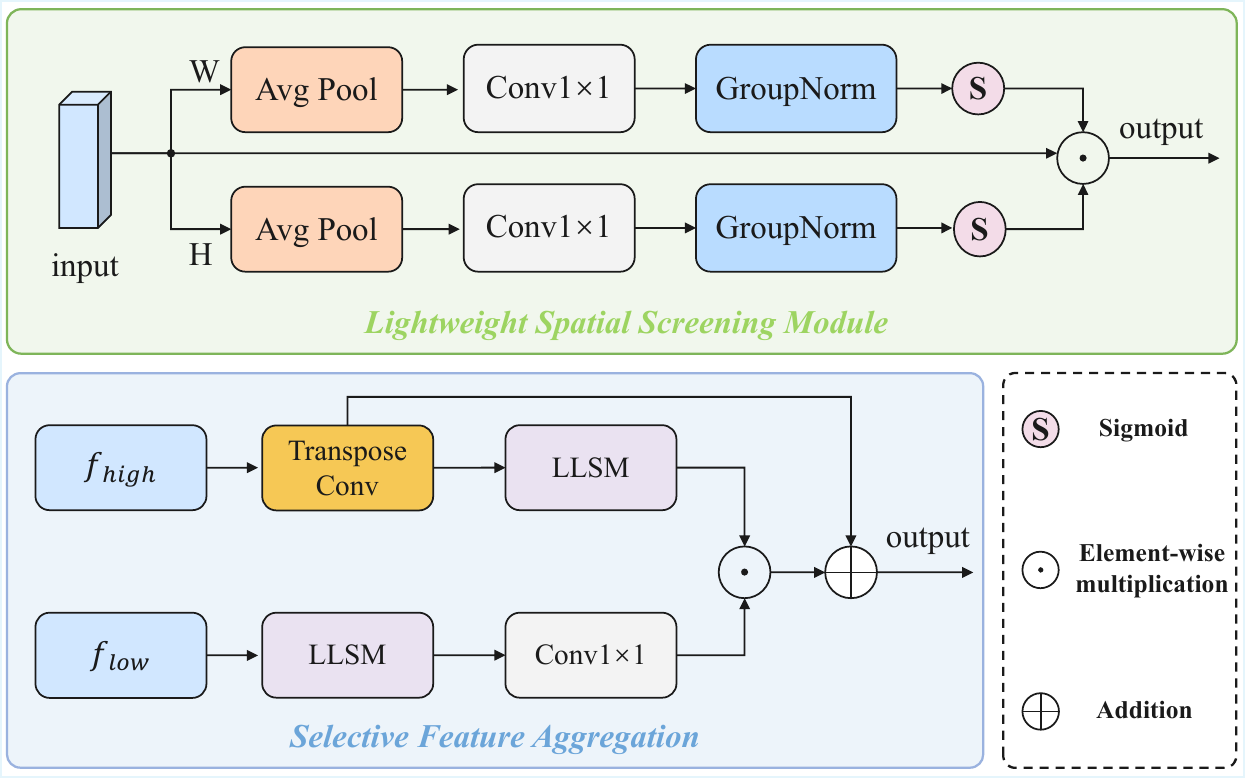}}
  \captionsetup{skip=0pt}
	\caption{The detailed structure of the adaptive screening pyramid network consists of a lightweight spatial screening module and a selective feature aggregation.}  
	\label{fig:Adaptive Screening Pyramid Network} 
	\vspace{-1.5em}
\end{figure}

In the PCB defect dataset, the multi-scale characteristics of defects pose challenges to the detection task and make the recognition ability of the model more complicated. This problem mainly stems from the different sizes of defects of different categories. Even for the same type of defects, their size and shape may vary significantly under different production processes and imaging equipment.
To address the inherent multi-scale challenges in PCB defect datasets, we designed an Adaptive Screening Pyramid Network  to achieve efficient multi-scale feature fusion, thereby improving the model's ability to identify defects of different sizes. As illustrated in Fig. \ref{fig:Adaptive Screening Pyramid Network}, ASPN consists of two core modules: (1) Lightweight Spatial Screening Module (LSSM); (2) Selective Feature Aggregation (SFA). LSSM refines features from different levels and removes redundant information. Subsequently, through the SFA, features of different scales are hierarchically integrated to enhance the model's fine-grained perception of PCB defects. This fusion method can effectively preserve the local details and overall structural information of the defects, thereby improving the robustness and accuracy of detection.

\subsubsection{Lightweight Spatial Screening Module}
LSSM improves the spatial perception ability of the model by effectively extracting key information and removing redundant features. This module can adaptively extract key information in the horizontal and vertical directions. Given an input feature map $\boldsymbol{\mathrm{X}}\in\mathbb{R}^{C\times H\times W}$, global context information is extracted through adaptive pooling operations in both height and width dimensions:
\begin{equation}
\boldsymbol{\mathrm{X}}_{\mathrm{H}}
=\sigma(Con\nu_{1\times1}(\mathrm{Pool}_h(\boldsymbol{\mathrm{X}}
))),
\end{equation}
\begin{equation}
\boldsymbol{\mathrm{X}}_{\mathrm{W}}
=\sigma(Con\nu_{1\times1}(\mathrm{Pool}_w(\boldsymbol{\mathrm{X}}
))),
\end{equation}
where ${Pool}_{h}$ and ${Pool}_{w}$ denote the adaptive average pooling operations applied along the height and width dimensions, respectively. The resulting pooled feature vector is then processed through a one-dimensional convolution layer to extract feature representations that are sensitive to spatial direction. Subsequently, the output is normalized using the Sigmoid function to generate channel-wise attention weights. These refined feature maps are fused with the original input features through element-wise multiplication, enabling effective integration of context-aware information into the network:
\begin{equation}
Y=\boldsymbol{\mathrm{X}}
\odot \boldsymbol{\mathrm{X}}_{\mathrm{H}}
\odot \boldsymbol{\mathrm{X}}_{\mathrm{W}}
.
\end{equation}

LSSM can enhance the ability to extract structured patterns in PCB defect images by decomposing feature selection into independent processing in the horizontal and vertical directions. Compared with traditional high-dimensional convolution operations, LSSM uses lightweight one-dimensional convolution to reduce computational overhead while maintaining strong feature expression capabilities. By selectively enhancing key features, the distinction of defect areas is improved, providing more accurate information for subsequent feature fusion and classification detection.

\subsubsection{Selective Feature Aggregation} 
SFA cleverly uses high-level features as weights to filter the basic semantic information in low-level features, thereby achieving strategic fusion of features. Given an input high-level feature $\boldsymbol{\mathrm{X}}_\mathrm{high}
\in\mathbb{R}^{C\times H\times W}$ and a low-level feature $\boldsymbol{\mathrm{X}}_\mathrm{low}
\in\mathbb{R}^{C\times H_1\times W_1}$, The high-level features are upsampled by a 3$\times$3 transposed convolution to match the size of the low-level features:
\begin{equation}
\boldsymbol{\mathrm{X}}_\mathrm{{high}}^\mathrm{T}
 = {TransposeConv}(\boldsymbol{\mathrm{X}}_\mathrm{{high}}
).
\end{equation}

The LSSM is applied to both high-level and low-level features separately to eliminate redundant information and extract salient semantic features. Subsequently, a 1$\times$1 convolution is employed to align the channel dimensions of the filtered low-level features. The filtered high-level features are then used as attention weights to perform point-wise multiplication with the dimension-adjusted low-level features. Finally, the resulting feature map is concatenated with the output of the transposed convolution to preserve more detailed information from earlier layers.

\begin{equation}
\boldsymbol{\mathrm{X}}_{\mathrm{low}}'=
\sigma(Con\nu_{1\times1}(\mathrm{LSSM}(\boldsymbol{\mathrm{X}}_{\mathrm{low}}
))),
\end{equation}
\begin{equation}
\boldsymbol{\mathrm{X}}_\mathrm{out}
=(\mathrm{LSSM}(\boldsymbol{\mathrm{X}}_\mathrm{{high}}^\mathrm{T})
\odot\boldsymbol{\mathrm{X}}_{\mathrm{low}}') +\boldsymbol{\mathrm{X}}_\mathrm{{high}}^\mathrm{T}.
\end{equation}

This fusion strategy effectively combines the global semantic information of high-level features and the spatial detail information of low-level features, enhancing the network's ability to express target features while avoiding the information redundancy problem caused by direct addition. The SFA module can significantly improve the effectiveness of feature fusion and performs well in PCB defect detection tasks.

\section{Experiments}
\subsection{Datasets} 
\begin{figure}[!tp]  
	\centerline{\includegraphics[page=1,trim = 0mm 0mm 0mm 0mm, clip, width=1\linewidth]{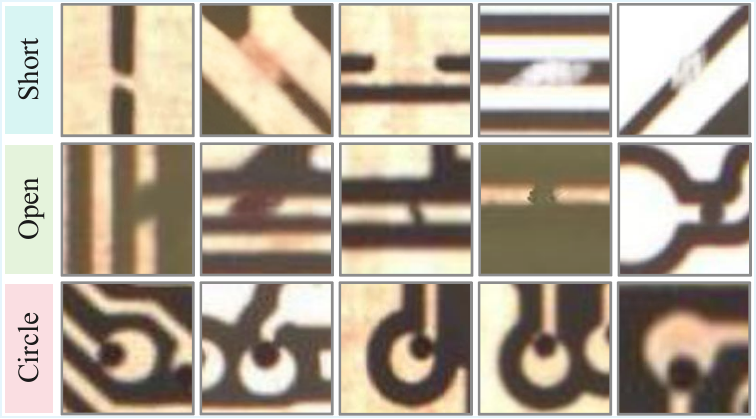}}
  \captionsetup{skip=0pt}
	\caption{Examples of typical bare board defects in the AOI-BarePCB dataset.}  
	\label{fig:defect}
	\vspace{-1.5em}
\end{figure}

This study uses a self-made AOI-BarePCB dataset, which is collected by an Automatic Optical Inspection (AOI) system in an actual production environment and focuses on the defect detection task at the bare board stage in the PCB manufacturing process. Unlike existing public datasets, this dataset does not contain finished boards, but directly targets bare PCB boards that have not been soldered and component mounted, which can better meet the quality inspection needs in actual production. The image resolution of the dataset is 640$\times$640, and the acquisition process uses industrial-grade AOI equipment to ensure the clarity of the image and the identifiability of the defects.  As illustrated in Fig. \ref{fig:defect}, the dataset contains three typical bare board defects of PCB: short circuit, open circuit and hole deviation. The detailed information are provided in  Tab. \ref{tab:Defect table}. The significance of the proposed AOI-BarePCB dataset lies in addressing the limitations of existing public datasets, which primarily focus on the finished board stage and often include soldering and component-mounting information. These datasets fail to accurately reflect the characteristics of early-stage defects in the manufacturing process, thus limiting the generalizability of defect detection algorithms on real production lines. In contrast, the AOI-BarePCB dataset is collected from the very source of industrial production, specifically targeting typical bare board defects such as short circuit, open circuit, and hole deviation. It fills the gap of high-quality data at the bare board stage and facilitates the development of early-stage defect detection models, ultimately improving both the efficiency and accuracy of quality inspection in intelligent manufacturing. Each image has been manually annotated, and the annotation format uses a rectangular frame.Each image has been manually annotated, and the annotation format uses a rectangular frame. As shown in the table, there are 800 images in the dataset, and we use 80\% as the training set and 20\% as the validation set and test set.

\begin{table}[h]
\centering
\renewcommand\arraystretch{1.2}
\caption{AOI-BarePCB Specific Information.}
\label{tab:Defect table}
\begin{tabular}{ccccl}
\toprule
AOI-BarePCB & short & open & circle  & total\\
\midrule
No. of Images & 541 & 660 & 228  & 800\\
No. of Defect & 1867 & 1926 & 571  & 4364\\
\toprule
\end{tabular}
\end{table}

\subsection{Experimental Setup}

\noindent\textbf{Implementation Details.} 
The experimental environment in this study runs on an Ubuntu system, and the experimental setting uses an Intel Xeon E5-2686 v4 CPU and a 24G NVIDIA TITAN RTX as the hardware configuration. The dataset uses the AOI-BarePCB dataset, and the model uses a batch size of 16, 300 epochs, and a learning rate of 0.0001. All networks are tuned using the AdamW optimizer, where the momentum and weight decay are 0.9 and 0.0001, respectively.

\noindent\textbf{Evaluation Metrics.} 
In order to comprehensively evaluate the effectiveness of the model, the evaluation indicators used in our experiment include: Precision, Recall, mAP, FPS, FLOPs and Params. Among them, Precision, Recall and mAP are calculated as follows:
\begin{equation}
Precision(P)=\frac{TP}{TP+FP},
\end{equation}
\begin{equation}
Recall(R)=\frac{TP}{TP+FN},
\end{equation}
\begin{equation}
mAP=\frac{1}{N}\sum_{c=1}^N\int_0^1P(R)dR.
\end{equation}
where precision measures the ratio of correct predictions among all positive detections, while recall reflects the proportion of actual positives that are correctly identified. Mean Average Precision (mAP) evaluates detection accuracy at an IoU threshold of 0.5. Frames Per Second (FPS) indicates inference speed, showing how many images the model can process per second. FLOPs and Params represent computational cost and model size, respectively, which are key factors for deployment.

\subsection{Comparative Experiment}
In this study, we selected several representative object detection methods for comparative experiments to comprehensively evaluate the performance advantages of the proposed method in PCB defect detection tasks. Among them, RT-DETR is used as a baseline, which adopts an efficient attention mechanism and an end-to-end detection framework to improve the inference speed while maintaining high detection accuracy. However, this method still has certain limitations in feature extraction and multi-scale feature fusion, and it is difficult to fully cope with complex defect detection scenarios.

To further verify the effectiveness of the proposed method, we compared a variety of detection frameworks based on Transformer and CNN. Transformer-based methods include DETR\cite{DETR}, DAB-DETR\cite{DAB-DETR}, Deformable-DETR\cite{Deformable-DETR},  DINO\cite{DINO}, RT-DETR\cite{RT-DETR}, D-FINE\cite{D-FINE} and DEIM\cite{DEIM} . These methods use the self-attention mechanism to model global features and optimize convergence speed, target localization ability, and computational efficiency. However, although these methods have improved in accuracy, some methods still have the problems of large computational complexity and slow inference speed, making them difficult to be directly applied to industrial detection tasks. On the other hand, we selected YOLOv8\cite{YOLOv8} and YOLO11\cite{YOLO11} as representatives of CNN-based methods. They use lightweight network architecture and efficient feature fusion strategy to achieve a good balance between detection accuracy and inference speed. In particular, YOLO11 combines the Transformer structure with the improved Anchor-Free mechanism to further improve detection accuracy and optimize inference efficiency, making it more suitable for real-time detection tasks. As shown in Tab. \ref{tab:Comparative experiment}, our model demonstrates clear advantages. It achieves an mAP of 0.956, outperforming all comparison methods, with a Precision of 0.937 and Recall of 0.941. The inference speed reaches 117.1 FPS, and the model remains lightweight with only 17.0M parameters and 48.2G FLOPs, showing an excellent balance between accuracy and efficiency.

\begin{figure*}[!tp]  
	\centerline{\includegraphics[page=1,trim = 0mm 0mm 0mm 0mm, clip, width=1\linewidth]{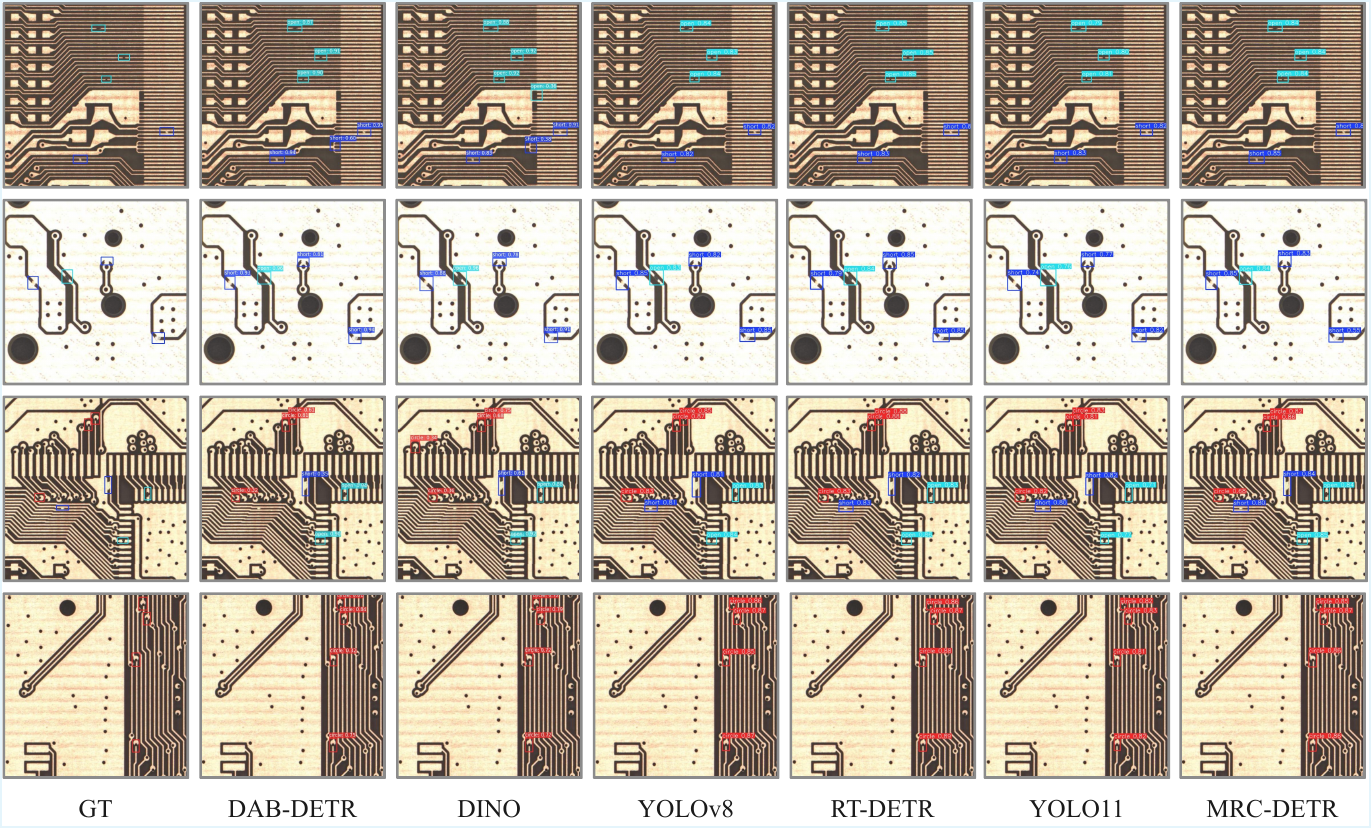}}
  \captionsetup{skip=0pt}
	\caption{Visualization of detection results across different models. For better clarity, results are best viewed in enlarged form.}  
	\label{fig:prediction}
	\vspace{-1.5em}
\end{figure*}

Although overall detection performance is significantly improved in terms of recall, inference speed, and model complexity, a slight decline in precision is observed. This may be attributed to the enhanced sensitivity of MRDC to low-confidence regions, which improves recall but slightly increases false positives. Additionally, the lightweight design of ASPN might reduce feature discriminability during multi-scale fusion, affecting fine-grained precision. However, this trade-off results in a more practical model that better meets the demands of real-time and resource-constrained industrial inspection scenarios. As shown in Fig. \ref{fig:prediction}, the visual comparison further highlights the superior detection performance of our proposed model under various challenging cases.



\captionsetup[table]{labelfont=bf, textfont=normalfont}
\begin{table*}[h]
\centering
\renewcommand\arraystretch{1.15}
\caption{Detection performance of different methods. The top three results are marked with \textcolor{red}{{red}}, \textcolor{blue}{{blue}}, and \textcolor{green}{{green}}, respectively.}
\label{tab:Comparative experiment}
\begin{tabular}{l|c|cccc|cc}  

\toprule
\toprule[0.5pt]

Model & Source & Precision$\uparrow$
& Recall$\uparrow$
& mAP$\uparrow$
& FPS$\uparrow$
& FLOPs (G)$\downarrow$
& Param. (M)$\downarrow$
\\

\midrule
DETR\cite{DETR}& 20'ECCV& 0.893 & 0.919 & 0.913 & 40.4 & 60.5 & 41.5 
\\

Deformable-DETR\cite{Deformable-DETR}&   21'ICLR & 0.924 & 0.919 & 0.945 & 19.8 & 125.9 & 40.1 
\\

DAB-DETR\cite{DAB-DETR}& 22'ICLR& 0.867& 0.880& 0.897& 25.0& {\color{green} 65.3}&43.7
\\

DINO\cite{DINO}& 22'CVPR& 0.913 & 0.901 & 0.942 & 14.1 & 183.3 & 47.5 
\\

YOLOv8\cite{YOLOv8}& 23'Ultralytics& {\color{green} 0.933}& 0.935 & {\color{green} 0.952}& {\color{red} 123.6}& 78.7 &  25.8\\

RT-DETR\cite{RT-DETR}& 24'CVPR& {\color{red} 0.948}& {\color{green} 0.939}& 0.947& 60.5& 129.6&42.0
\\

YOLO11\cite{YOLO11}& 24'Ultralytics& 0.931 & {\color{blue} 0.940}& {\color{blue} 0.953}& {\color{blue} 123.3}&  67.7& {\color{green} 20.0}\\
 D-FINE\cite{D-FINE}& 25'ICLR& 0.928& 0.934& 0.943& 94.1& {\color{blue} 56.3}&{\color{blue} 19.1}\\
 DEIM\cite{DEIM}& 25'CVPR& 0.932& 0.937& 0.950& 73.9& 92.5&31.3\\

\rowcolor{my_color}
MRC-DETR & -& {\color{blue} 0.937}& {\color{red} 0.941}& {\color{red} 0.956}& {\color{green} 117.1}& {\color{red} 48.2}& {\color{red} 17.0}\\

\bottomrule
\end{tabular}
\end{table*}

\subsection{Ablation Study} %
In order to comprehensively evaluate the contribution of each key module in the proposed method to  detection performance, this study conducted ablation experiments based on RT-DETR, focusing on analyzing the impact of MRDC and ASPN. The results of the ablation experiment are shown in Tab. \ref{tab:Ablation Experiment}, where setting (a) denotes the use of MRDC only, setting (b) uses ASPN only, and setting (c) incorporates both MRDC and ASPN. 

The experimental results show that in setting (a), after introducing MRDC, the mAP increased from 0.947 to 0.949, indicating that the module effectively optimizes the feature extraction process and enhances detection accuracy. Meanwhile, the number of parameters decreased from 42.0M to 36.6M, and the FLOPs decreased from 129.6G to 118.5G, demonstrating that MRDC improves precision while reducing computational cost. However, the recall dropped from 0.939 to 0.931, which may be attributed to the change in feature update strategy, possibly leading to missed detections of certain boundary objects.

In setting (b), the introduction of ASPN further reduced computational complexity, with the number of parameters dropping to 19.9M and the FLOPs reduced to 53.7G, a 58. 6\% decrease compared to the baseline, while maintaining an mAP of 0.946. This indicates that ASPN significantly reduces the computational burden without severely compromising detection accuracy. However, the recall declined to 0.922, probably due to the reduced feature representation capacity that affected the detection of some targets.

\begin{figure}[!tp]  
	\centerline{\includegraphics[page=1,trim = 0mm 0mm 0mm 0mm, clip, width=1\linewidth]{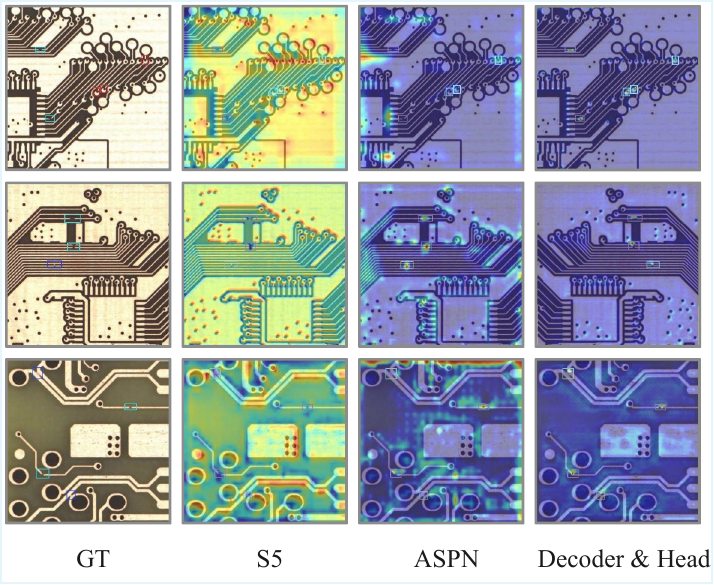}}
  \captionsetup{skip=0pt}
	\caption{Visualization of detection heatmaps produced by the proposed model.}    
	\label{fig:heatmap}
	\vspace{-1.5em}
\end{figure}

After combining MRDC and ASPN in setting (c), the model achieved an mAP of 0.956, surpassing the baseline by 0.9 percentage points. Meanwhile, the parameter count decreased to 17.0M and the FLOPs fell to 48.2G, reflecting a simultaneous improvement in both accuracy and computational efficiency. The recall rate also increased  to 0.941, indicating that the combined approach effectively compensates for the individual limitations of MRDC and ASPN. To further demonstrate the ability of the model to localize defects, heatmap visualizations of detection results are presented in Fig. \ref{fig:heatmap}. These results confirm the suitability of the proposed method for lightweight industrial defect detection applications.

\begin{table*}[h]
\centering
\renewcommand\arraystretch{1.15}
\caption{Ablation study on the effectiveness of MRDC and ASPN modules.}
\label{tab:Ablation Experiment}
\begin{tabular}{ccccccccc}\toprule

Index & MRDC & ASPN & Precision$\uparrow$& Recall$\uparrow$& mAP$\uparrow$& FPS$\uparrow$& FLOPs (G)$\downarrow$& Param. (M)$\downarrow$\\\midrule

baseline & \ding{55} & \ding{55} & \textbf{0.948}& \underline{0.939}& 0.947 & 60.5 & 129.6 & 42.0 \\
a & \ding{51} & \ding{55} & \underline{0.946}& 0.931 & \underline{0.949}& 50.1 & 118.5 & 36.6 \\
b & \ding{55} & \ding{51} & 0.928 & 0.922 & 0.946 & \textbf{152.8}& \underline{53.7}& \underline{19.9}\\
c & \ding{51} & \ding{51} & 0.937& \textbf{0.941}& \textbf{0.956}& \underline{117.1}& \textbf{48.2}& \textbf{17.0}\\ \bottomrule

\end{tabular}
\end{table*}

\begin{table*}[h]
\centering
\renewcommand\arraystretch{1.2}
\caption{Comparison of different lightweight attention modules in ASPN.}
\label{tab:Comparison of different lightweight attention modules in ASPN}
\begin{tabular}{cccccc}
\toprule
Attention Methods& Precision$\uparrow$& Recall$\uparrow$& mAP$\uparrow$& FLOPs (G)$\downarrow$ &Param. (M)$\downarrow$\\
\midrule
+SE\cite{SE}& 0.912& 0.901& 0.939&   49.3& \textbf{15.1}\\
+SGE\cite{SpatialGroupEnhance}& 0.928& 0.926& 0.944& 49.0& 16.1\\

 +CAA\cite{CAA}& 0.926& 0.930& 0.948& 52.2&15.7\\
 +LSSM& \textbf{0.937}& \textbf{0.941}& \textbf{0.956}& \textbf{48.2}&17.0\\ \bottomrule
\end{tabular}
\end{table*}

To further evaluate the effectiveness of the proposed Lightweight Spatial Screening Module (LSSM) module in ASPN, we conducted a set of controlled ablation experiments. In these experiments, the LSSM component was individually replaced by three widely-used lightweight attention mechanisms: Squeeze-and-Excitation (SE)\cite{SE}, Spatial Group Enhancement (SGE)\cite{SpatialGroupEnhance}, and Context Anchor Attention (CAA)\cite{CAA}, while keeping the rest of the ASPN architecture unchanged.

As shown in Tab.  \ref{tab:Comparison of different lightweight attention modules in ASPN}, although the number of model parameters is slightly reduced by about 1-2M after using SE, SGE and CAA, the precision, recall and mAP indicators are not better than the original LSSM solution. In particular, while keeping the model lightweight, LSSM shows better feature enhancement capabilities, further improving the detection performance. In addition, efficiency indicators such as FLOPs also show that LSSM has a good balance between speed and computational efficiency.

These results validate the superiority of LSSM as the core component in ASPN and demonstrate that its design better suits the lightweight yet accurate requirements of PCB defect detection.

\section{Conclusion}

In this paper, we systematically review the current challenges in PCB defect detection and propose an enhanced Transformer-based object detection framework, termed MRC-DETR, specifically designed for bare PCB defect inspection. To address the issues of insufficient feature representation and computational redundancy, we introduce the Multi-Residuals Directional Coupled Block, which enhances the model’s adaptive feature learning capability while reducing computational overhead, thereby improving both accuracy and efficiency. Furthermore, to mitigate the excessive computation introduced by cross-layer information interaction, we propose the Adaptive Screening Pyramid Network, which accurately localizes regions of interest and effectively selects critical information from low-level feature maps that are relevant to high-level semantic features, enabling efficient multi-scale feature fusion. In addition, a dedicated dataset tailored for bare PCB defect detection is constructed to provide more realistic and comprehensive data support, facilitating the practical deployment and optimization of defect detection algorithms in industrial scenarios. Compared with existing PCB defect detection approaches, MRC-DETR achieves superior performance, as demonstrated through extensive qualitative and quantitative evaluations.


%
%

\bibliographystyle{IEEEtran}
\bibliography{IEEEabrv,UniUIR}

\end{document}